\documentclass{article}

\usepackage{microtype}
\usepackage{graphicx}
\usepackage{subfigure}
\usepackage{booktabs} %
\usepackage{times}
\usepackage{epsfig}
\usepackage{graphicx}
\usepackage{amsmath}
\usepackage{amssymb}
\usepackage{float}
\usepackage{subfigure}
\usepackage{makecell}
\usepackage{multirow}
\usepackage{amsfonts}

\usepackage{hyperref}

\usepackage[accepted]{icml2019}

\icmltitlerunning{Understanding Adversarial Robustness Through Loss Landscape Geometries}

\begin{document}

\onecolumn
\icmltitle{Understanding Adversarial Robustness Through Loss Landscape Geometries}

\icmlsetsymbol{equal}{*}

\begin{icmlauthorlist}
\icmlauthor{Vinay Uday Prabhu*}{unifyid}
\icmlauthor{Joyce Xu*}{stanford,intern}
\icmlauthor{Dian Ang Yap*}{stanford,intern}
\icmlauthor{John Whaley}{unifyid}
\end{icmlauthorlist}

\icmlaffiliation{stanford}{Department of Computer Science, Stanford University, CA 94305, USA}
\icmlaffiliation{unifyid}{UnifyID, Redwood City, CA 94063, USA}
\icmlaffiliation{intern}{Work done as as intern/fellow at UnifyID}

\icmlcorrespondingauthor{Joyce Xu}{jexu@stanford.edu}
\icmlcorrespondingauthor{Dian Ang Yap}{dayap@stanford.edu}

\icmlkeywords{Machine Learning, ICML, Adversarial Detection, Robustness, Loss Landscapes, Adversarial Augmentation}

\vskip 0.3in

\printAffiliationsAndNotice{\icmlEqualContribution} %

\begin{abstract}
The pursuit of explaining and improving generalization in deep learning has elicited efforts both in regularization techniques as well as visualization techniques of the loss surface geometry. The latter is related to the intuition prevalent in the community that flatter local optima leads to lower generalization error. In this paper, we harness the state-of-the-art ``filter normalization'' technique of loss-surface visualization to qualitatively understand the consequences of using adversarial training data augmentation as the explicit regularization technique of choice. Much to our surprise, we discover that this oft deployed adversarial augmentation technique does not actually result in ``flatter'' loss-landscapes, which requires rethinking adversarial training generalization, and the relationship between generalization and loss landscapes geometries.
\end{abstract}
\label{submission}

\section{Introduction}
State of the art deep learning models that exhibit low generalization error intriguingly exist in the regime where the number of trainable parameters in the model is often greater than the number of training data points. In order to address the susceptibility of such models to over-fitting, there exists a formidable body of literature of regularization techniques, which broadly fall into two categories. The first category is \textit{implicit regularization} \cite{neyshabur2017implicit} that includes examples such as early stopping during training or small-norm solution inducing Stochastic Gradient Descent (SGD). The second category is \textit{explicit regularization} that includes techniques such as using $\ell_1$ or $ \ell_2$-norm weight penalty, training data augmentation, architectural changes (such as introducing skip connections) and dropouts. Training data augmentation based explicit regularization in turn falls into two categories. The first entails usage of \textit{classical} hand-crafted transformations such as elastic transformation, random cropping, perturbations of image attributes such as brightness and contrast, and synthetic oversampling. The second entails harnessing adversarial perturbations to augment the dataset \cite{kurakin2016adversarial}. The work presented in this paper fits squarely into this second category, that is also termed \textit{adversarial training}.

Co-temporal to this evolution of regularization techniques for deep learning, has been the quest to both qualitatively and quantitatively investigate the nexus between the geometry of the loss-landscape (flatness of the local optima) and  the ability of regularized-model to generalize well \cite{li2018visualizing}. Combining these two potent and hitherto unconnected set of ideas, in this paper, we seek to qualitatively address the following question:

\vskip 0.1in
{\centering
  \textit{``What happens to the loss surface of deep-nets
  \\when we use explicit regularization using adversarial data augmentation?"}\par
}
\vskip 0.1in

Guided by the pre-eminent merits of adversarial training, our \textit{ansatz} at the beginning of this experimental work was that we would encounter mild to impressive \textit{flattening} of the loss surfaces. Much to our surprise, our findings did not meet this expectation, which requires rethinking generalization \cite{zhang2016understanding} in the context of adversarial training. Our work also provides an experimental approach from other work with theoretical perspectives on the notion of adversarial training \cite{raghunathan2019adversarial}.

Through this dissemination, we'd like to call upon the community at large to further investigate this finding and have duly open sourced all our code\footnote{\url{https://github.com/joycex99/adversarial-training}} in order to accelerate reproducibility and experimentation. This is a work in progress and we are actively updating the repository with more experiments.

\section{Methods and Experiments}
\subsection{Adversarial Attacks}
We experiment with adversarially augmented training data using three different common attacks, with all three attacks using the $\ell_\infty$ metric. The first is the Fast Gradient Sign Method (FGSM) as an $\ell_\infty$-bounded adversary, which computes an adversarial sample $x_{\text{adv}}$ from $x$ \cite{goodfellow2014explaining} by modifying the input data to maximize the loss based on the backpropagated gradients with some perturbation $\epsilon$, which can be expressed as:
\begin{equation}
    x_{\text{adv}} = x + \epsilon \text{sgn}\left(\nabla_x L(\theta, x, y)\right).
\end{equation}
Secondly, we consider projected gradient descent (PGD) as a more powerful adversary, which is considered a ``universal" first-order technique and is essentially a multi-step variant $\text{FGSM}^k$ with random starts \cite{kurakin2016adversarial}\cite{madry2017towards}:
\begin{equation}
    x_{\text{adv}}^{t+1} = \text{Clip}\left(x^t + \alpha \text{sgn}\left(\nabla_x L(\theta, x, y)\right)\right).
\end{equation}
Thirdly, we use the Spatially Transformed Adversarial Attack (stAdv) that crafts more perceptually realistic attacks by modifying image \textit{geometry} \cite{xiao2018spatially} and changing the positions of pixels. Instead of directly changing pixel values, these adversaries optimize the amount of displacement in each dimension, i.e. the flow vector $f_i = (\Delta u^{(i)}, \Delta v^{(i)}$). Each pixel value from input location $(u^{(i)}, v^{(i)})$ corresponds to a flow-displaced pixel in the adversarial image:
\begin{equation}
    (u^{(i)}, v^{(i)}) = \left(u_{\text{adv}}^{(i)} + \Delta u^{(i)}, v_{\text{adv}}^{(i)} + \Delta v^{(i)}\right).
\end{equation}
Since these locations can be fractional coordinates, each  adversarial pixel $x_{\text{adv}}^{(i)}$ is calculated through bilinear interpolation of $N$ neighboring pixels:
\begin{equation}
    x_{\text{adv}}^{(i)} = \sum_{q \in N(u^{(i)}, v^{(i)})} x^{(q)}\left(1-|u^{(i)} - u^{(q)}|\right)\left(1-|v^{(i)} - v^{(q)}|\right).
\end{equation}
\subsection{Adversarial Training}
We experiment with a standard ResNet-32 model architecture \cite{he2016deep}. As a base model, we first train a clean model on the original CIFAR10 dataset until convergence. Then, for each attack, we generate a single (untargeted) adversarial example per training point in the original dataset, such that the new augmented dataset has a one-to-one ratio of clean and adversarial samples. The base model is then fine-tuned for 200 epochs on the augmented dataset, with the learnt weights then used for loss landscape visualizations.

We recognize and want to highlight that this choice of training procedure likely has an impact on the results we obtain. In particular, pretraining and then performing adversarial fine-tuning may result in a different loss landscape than performing adversarial training from scratch. We investigate this training procedure because we are interested in how much adversarial training can \textit{increase} robustness relative to existing trained models, potentially as part of a multi-step process to improve model generalization. However, we are also interested in and encourage future exploration of loss landscapes of models adversarially trained from scratch. 

\subsection{Loss Landscape Visualization}
Traditional 2D contour plots visualize the change in loss moving from some center point $\theta^*$ (e.g. the optimization minimum) out along any two direction vectors $\delta$ and $\eta$:
\begin{equation}
    f(\alpha, \beta) = L(\theta^* + \alpha\delta + \beta\eta)
\end{equation}
Since naively adding these direction vectors to $\theta^*$ (which can vary greatly in magnitude model to model) sacrifices the scale invariance of network weights, we normalize the direction vectors $d$ by the Frobenius norm of the filter:  
\begin{equation}
    d_{i,j} = \frac{d_{i,j}}{||d_{i,j}||} ||\theta_{i,j}||
\end{equation}
where $d_{i,j}$ is the $j$th filter of the $i$th layer of $d$ \cite{li2018visualizing}.

With this modification, even if the adversarial models converge upon different magnitudes of weights, the \textit{relative} curvatures and geometry of their loss landscapes are directly comparable.

\section{Results and Visualizations}

We evaluate original and augmented models on Top-1 accuracy on both the original test data and adversarial inputs. We use FGSM with the $\ell_{\infty}$ radius of $8/255$, PGD with radius $1/255$ for $10$ iters, and stAdv with radius $0.3/64$. The relative robustness of our adversarially-trained models supports accuracies as reported by existing literature \cite{xiao2018spatially}.

\begin{figure*}[!tb]
\centering     %
\subfigure[Un-augmented]{\label{fig:a}\includegraphics[width=70mm]{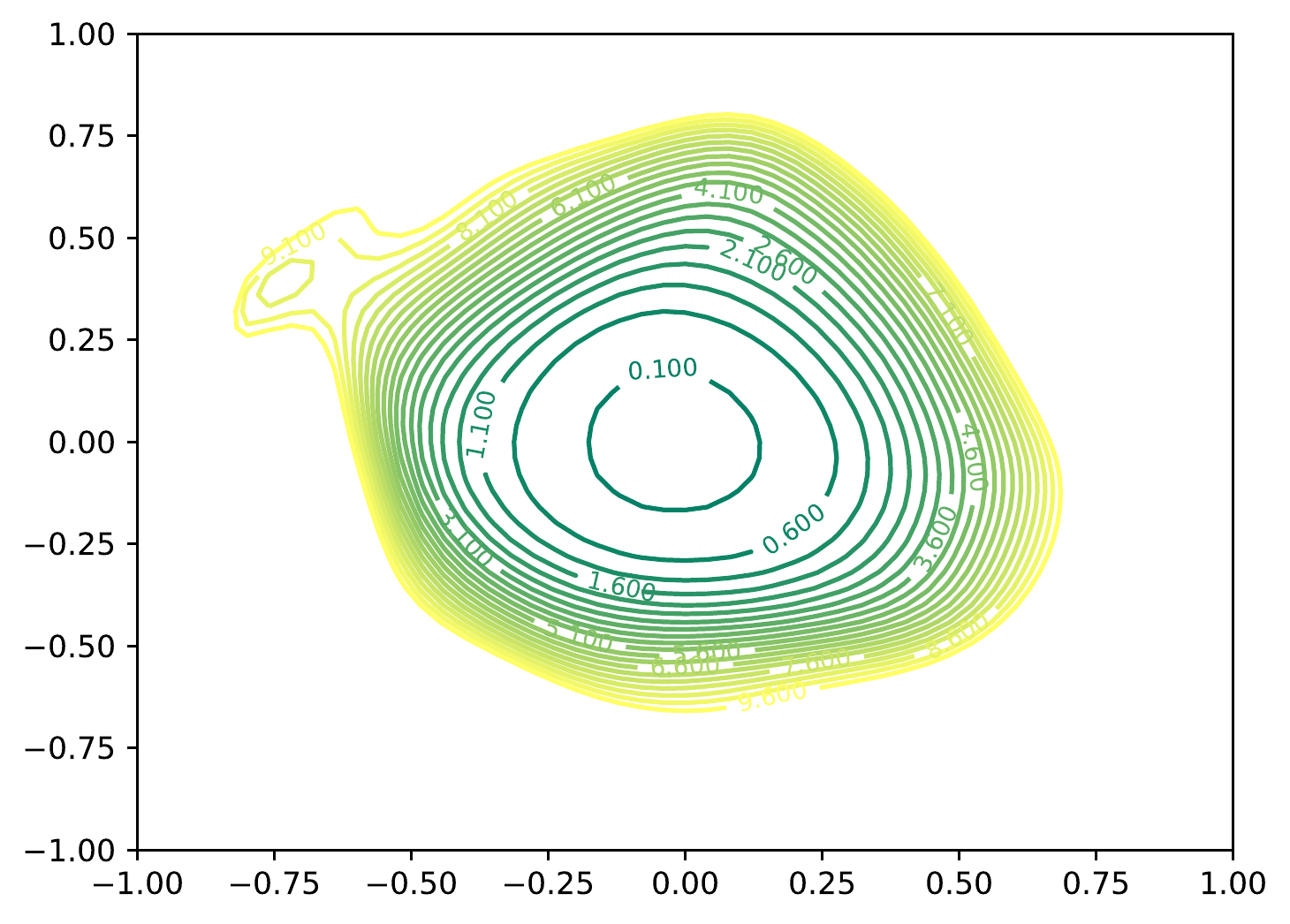}}
\subfigure[FGSM augmentation]{\label{fig:b}\includegraphics[width=70mm]{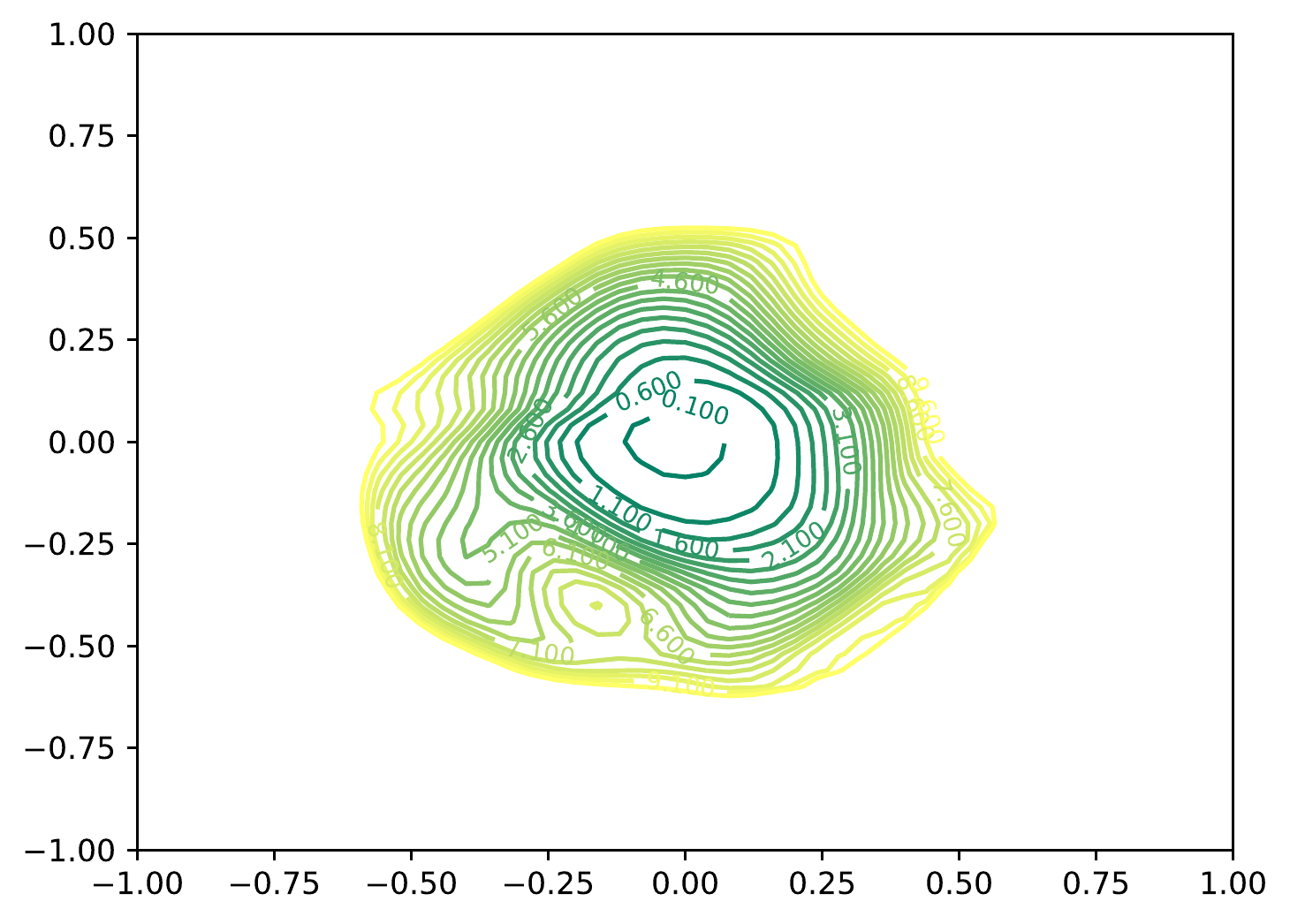}}
\subfigure[PGD augmentation]{\label{fig:c}\includegraphics[width=70mm]{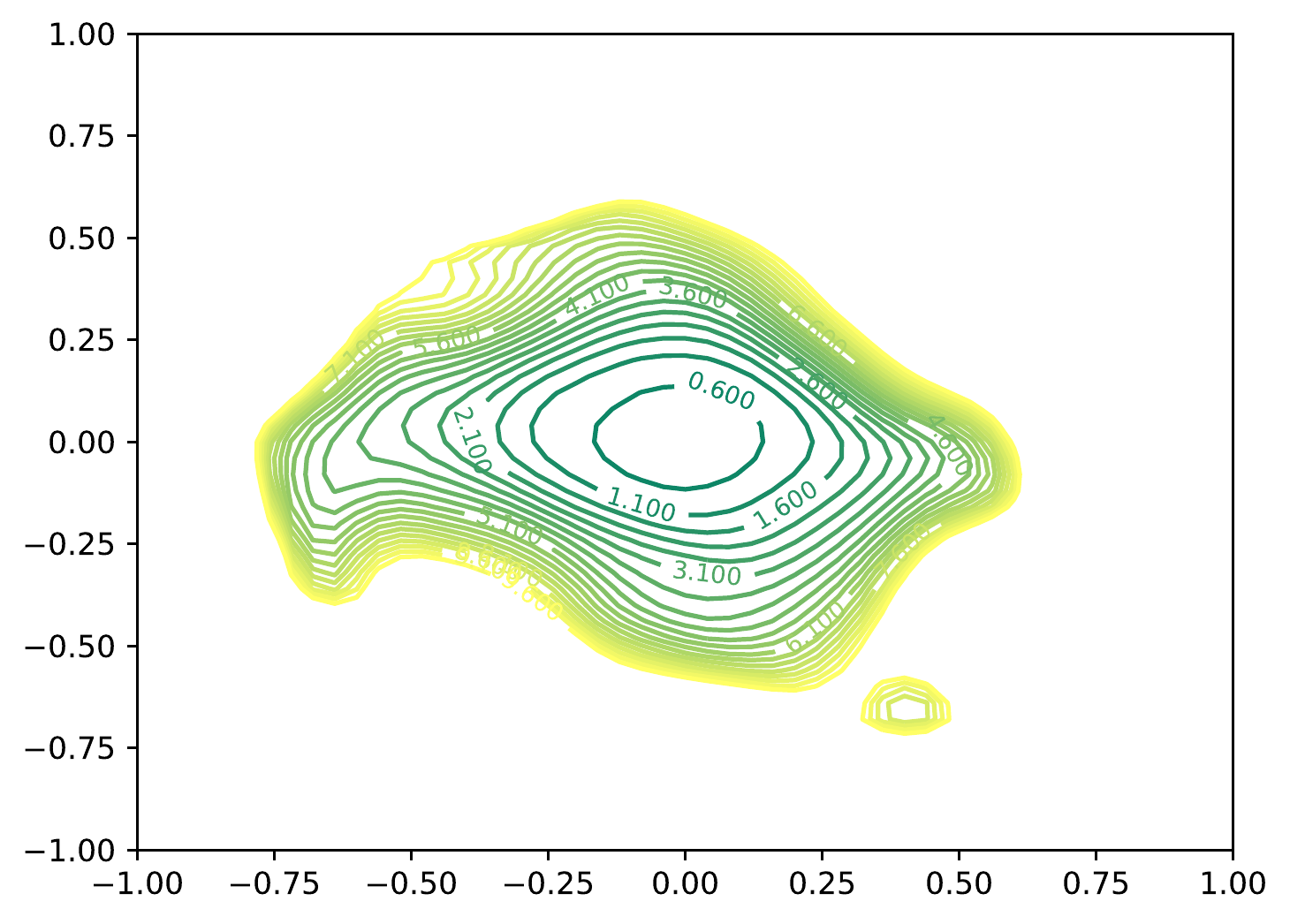}}
\subfigure[stAdv augmentation]{\label{fig:d}\includegraphics[width=70mm]{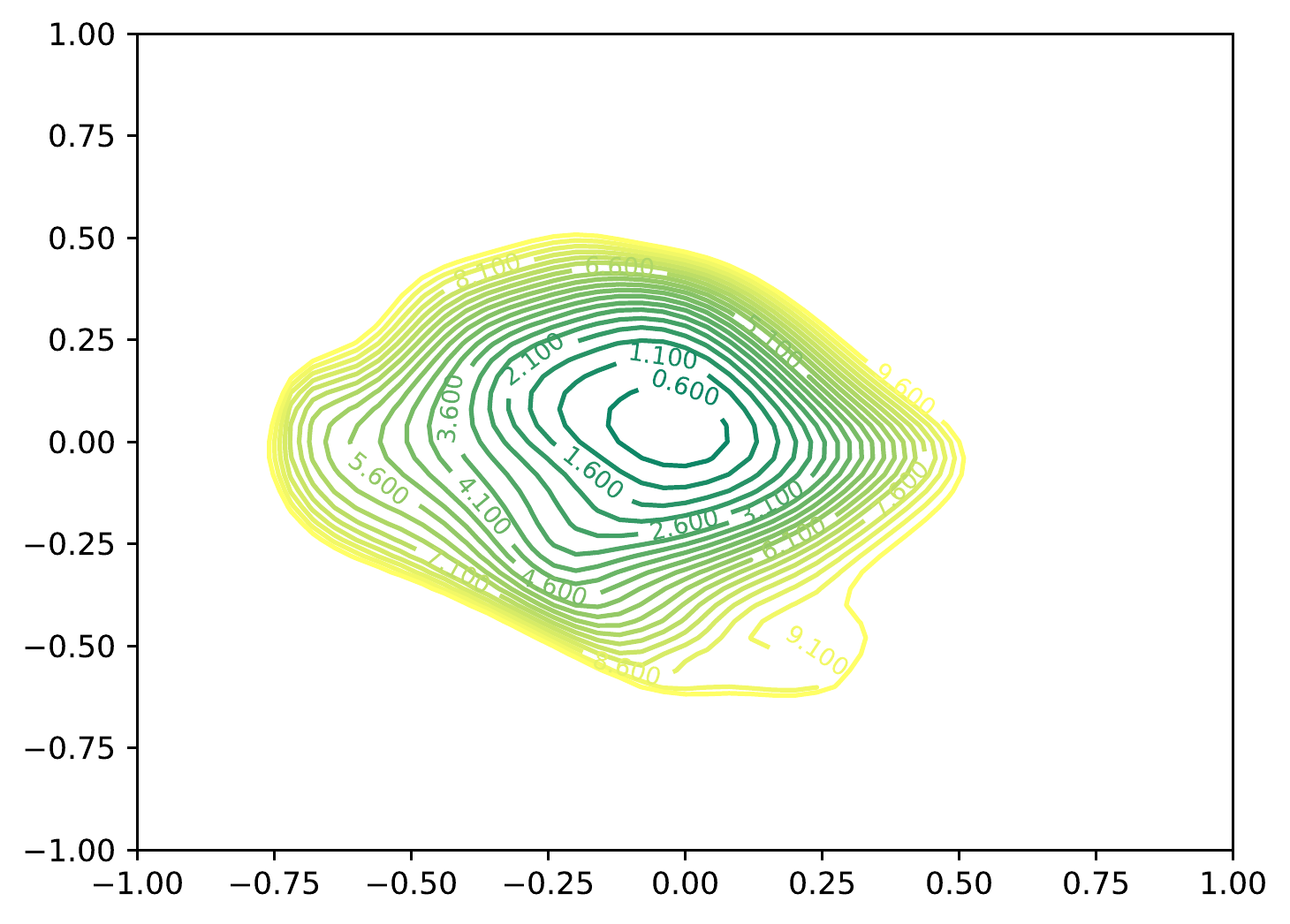}}
\caption{Visualizations of loss landscapes of the same model. (b)-(d) are losses fine-tuned augmented data from different attacks.}
\end{figure*}

\begin{table}[!tbp]
\caption{Top-1 Accuracies for Original and Adversarially-Augmented Data}
\begin{sc}
\begin{center}
\begin{tabular*}{0.5\columnwidth}{ccccc }
\hline
    \thead{Augmented \\ with} & \thead{Ground \\ Acc.} & \multicolumn{3}{c}{\thead{Adversarial \\ Acc.}}\\
    \cline{3-5}
    && FGSM & PGD & stAdv \\
\hline

None & \textbf{92.64} & 16.88 & 0.0 & 0.0 \\
FGSM & 90.11 & \textbf{75.23} & 12.11 & 0.0 \\
PGD & 87.34 & 60.52 & 51.20 & 13.85\\
stAdv & 91.05 & 55.84 & \textbf{52.98} & \textbf{25.49}\\
\hline
\end{tabular*}
\end{center}
\end{sc}
\end{table}

Surprisingly, the loss landscape geometries do not follow intuition of common practices. Since adversarial training has been shown to improve generalization, and models with low generalization error tend to have smoother loss landscapes, we originally expected adversarially-trained nets to have smoother loss landscapes, fewer local minima and greater convexity. However, the clean model (trained on the original data) appears to have the smoothest/most convex loss landscape (Fig \ref{fig:a}). Meanwhile, the FGSM-augmented model has the sharpest/least convex geometry, followed by PGD and then the comparatively smoother stAdv (Fig \ref{fig:b}, \ref{fig:c}, \ref{fig:d}), which supports the idea that adversarial training actually hurts generalization through through trade-offs between the \textit{standard accuracy} and \textit{adversarially robust accuracy} of a model

We also note that there is a positive correlation between the smoothness/convexity of the adversarial loss landscapes and the relative similarity between their perturbed images and the originals. Table 2 shows the average difference of perturbed images from the original for each attack, measured by 1 - SSIM (the Structural SIMilarity index) \cite{flynn2013image}\cite{wang2004image}. Our results demonstrate that smaller 1-SSIM differences correspond to smoother loss landscapes. Specifically, models trained on stAdv perturbations - which intentionally modify image geometry as opposed to arbitrary pixel values - have the smoothest adversarial loss landscape. Meanwhile, training on FGSM-perturbed images that have lower structural similarity to the original results in a particularly sharp and chaotic landscape. 

\begin{figure*}[!tbp]
\centering     %
\subfigure[Un-augmented]{\label{fig:2a}\includegraphics[width=83mm]{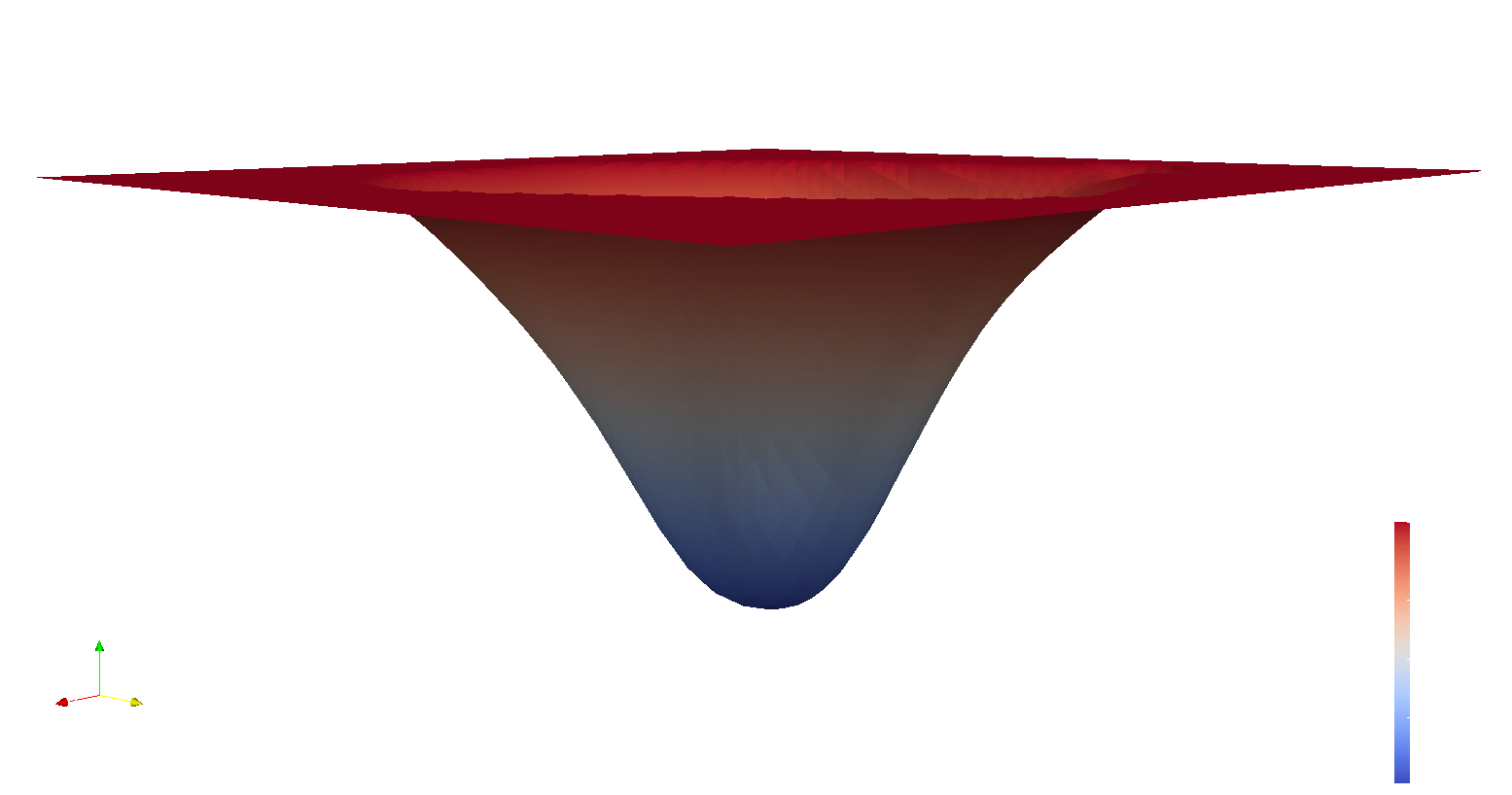}}
\subfigure[FGSM augmentation]{\label{fig:2b}\includegraphics[width=83mm]{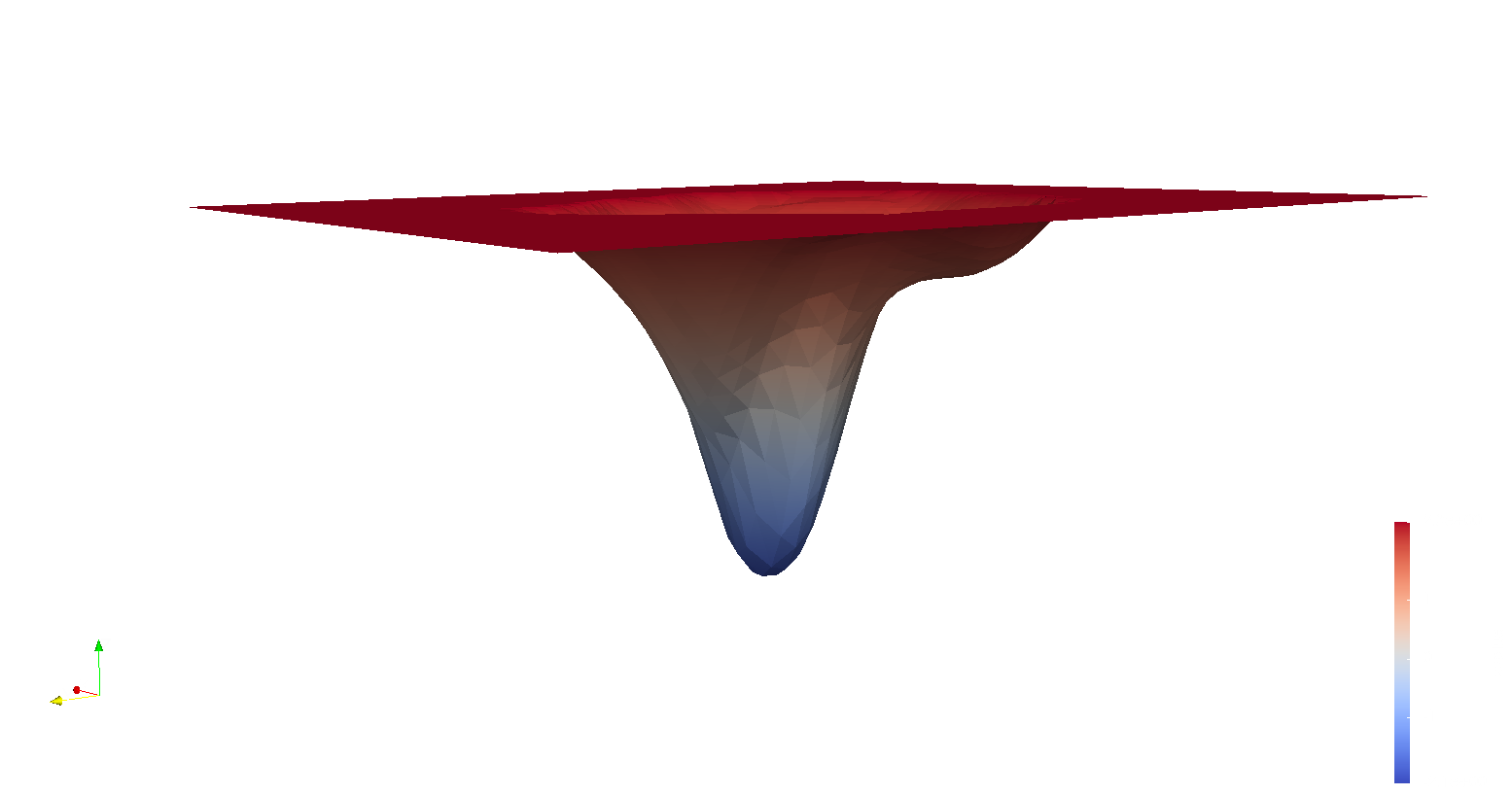}}
\subfigure[PGD augmentation]{\label{fig:2c}\includegraphics[width=83mm]{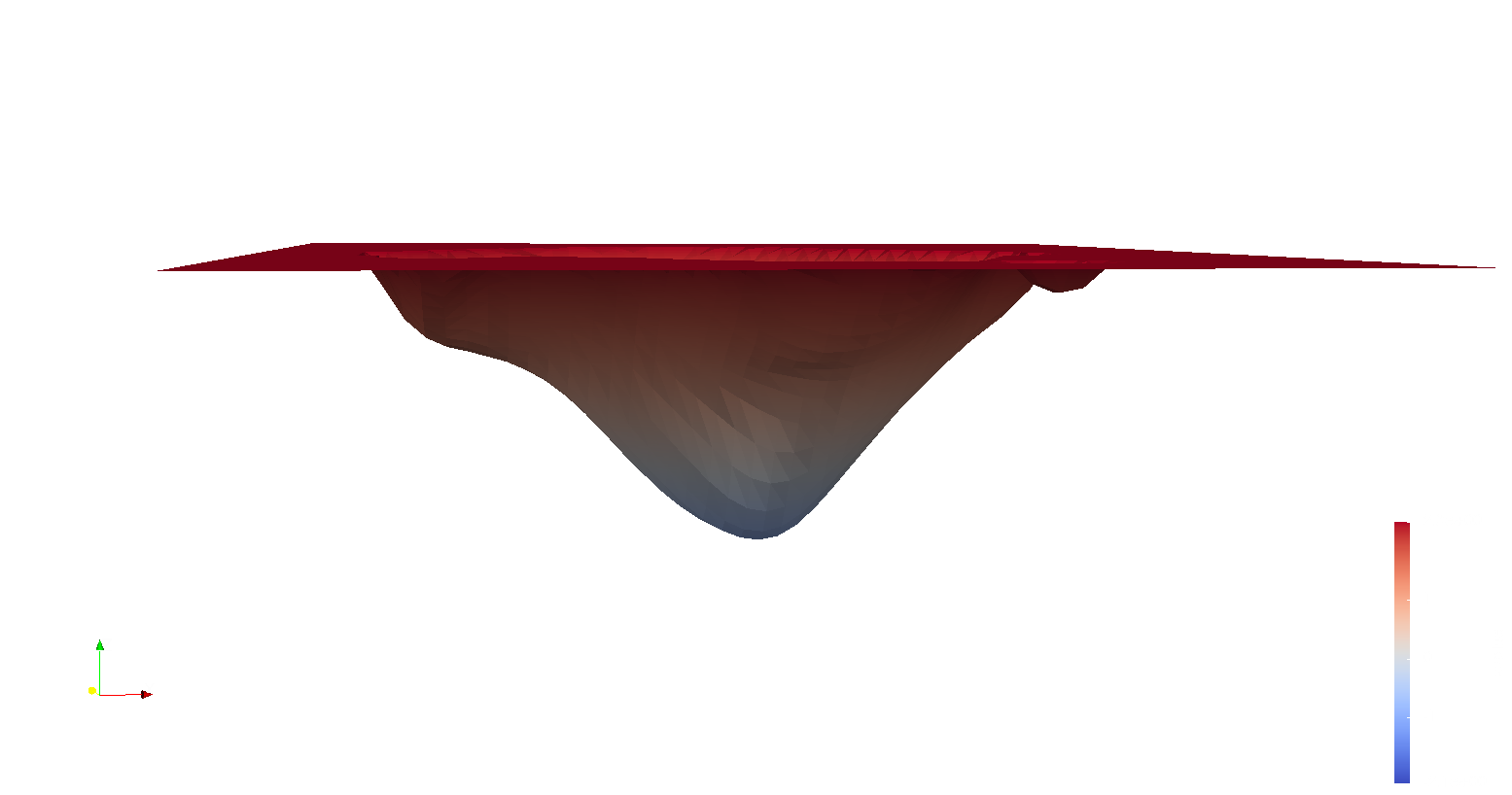}}
\subfigure[stAdv augmentation]{\label{fig:2d}\includegraphics[width=83mm]{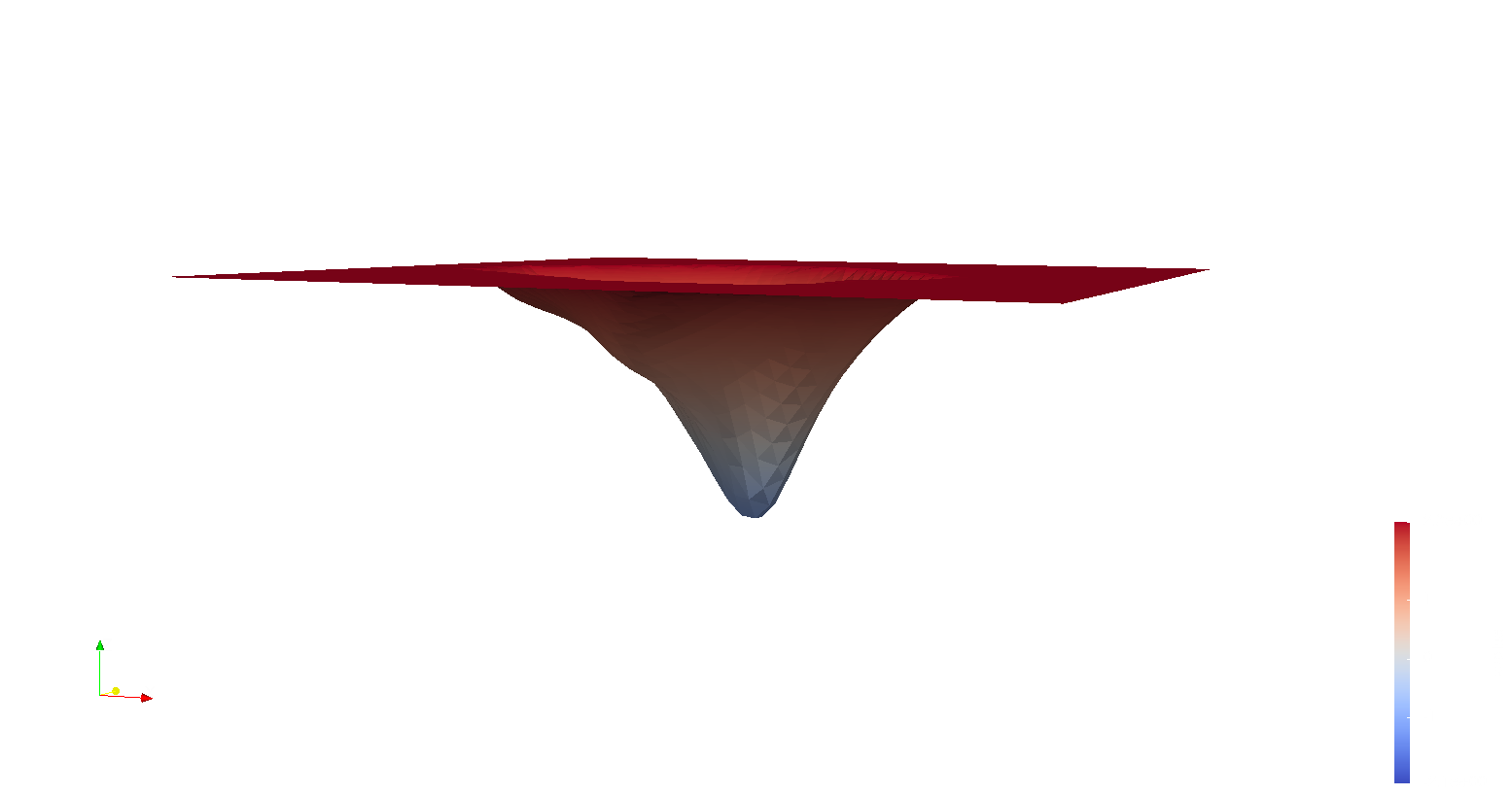}}
\caption{3D Visualizations of loss landscapes of the same model. (b)-(d) are losses fine-tuned augmented data from different attacks.}
\end{figure*}
\begin{table}[!htbp]
\caption{Average SSIM Distance of Adversary from Original}
\begin{sc}
\begin{center}
\begin{tabular}{cc}
\hline
Data Augmented with & 1-SSIM\\
\hline
FGSM & 0.049344 \\
PGD & 0.008206\\
stAdv & \textbf{0.002801}\\
\hline
\end{tabular}
\end{center}
\end{sc}
\end{table}

\section{Conclusion}
The loss landscape geometries of adversarially-trained neural networks reveal that overall model \textit{generalization} does not necessarily improve with resilience to certain attacks. The sharper loss landscapes of adversarially-trained models are not only an unintuitive, surprising result for ongoing model interpretability efforts, but an important point of consideration for developing new adversarial training techniques.

Fine-tuning models on augmented data may also contribute to sharper and more chaotic local minima, since the optimization path from former local minima can differ from the optimization path from randomly initialized weights. Once again, we encourage further investigation in comparing our results to those obtained from non-pretrained models. 

Based on our investigations, out of our three adversarial augmentations, stAdv shows the most promise in terms of accuracy and generalization abilities. Even so, under adversarial fine-tuning, we show models can default to memorizing perturbations and generalizing only \textit{based on structural image similarity} \cite{yuan2019adversarial}\cite{zhang2018adversarial}. It is possible that combining multiple attack methods might improve overall generalization and converge upon relatively flat loss landscapes, and is a potential avenue for future investigation. 

\newpage
\bibliography{example_paper}
\bibliographystyle{icml2019}

\end{document}